\documentclass[10pt,twocolumn,letterpaper]{article}
\usepackage[pagenumbers]{cvpr} 

\usepackage{multirow}
\usepackage{makecell}
\usepackage[dvipsnames]{xcolor}
\usepackage{comment}
\usepackage{pifont}
\usepackage{colortbl}
\usepackage{tabularx}
\usepackage{adjustbox}
\usepackage{mathrsfs}
\usepackage{xspace}

\newcommand{\mypara}[1]{\noindent\textbf{#1}}
\newcommand{\cmark}{\textcolor{ForestGreen}{\ding{51}}}%
\newcommand{\xmark}{\textcolor{red}{\ding{55}}}%

\usepackage{xspace}
\newcommand{\fm}{FM\xspace}
\newcommand{\ours}{JRM\xspace}

\definecolor{cvprblue}{rgb}{0.21,0.49,0.74}
\usepackage[pagebackref,breaklinks,colorlinks,allcolors=cvprblue]{hyperref}

\def\paperID{****} 
\def\confName{CVPR}
\def\confYear{2026}

\title{\ours: Joint Reconstruction Model for Multiple Objects without Alignment}

\begin{document}

\author{
    Qirui Wu$^{1,2,\dagger}$ \quad Yawar Siddiqui$^{1}$ \quad Duncan Frost$^{1}$ \quad Samir Aroudj$^{1}$ \quad Armen Avetisyan$^{1}$ \\
    Richard Newcombe$^{1}$ \quad Angel X. Chang$^{2}$ \quad Jakob Engel$^{1}$ \quad Henry Howard-Jenkins$^{1}$
    \\[2mm]
    $^{1}$Meta Reality Labs Research \qquad $^{2}$Simon Fraser University
}

\newcommand{\teaserfigure}{
\vspace{-2.em}
\begin{center}
\captionsetup{type=figure}
\includegraphics[width=\textwidth]{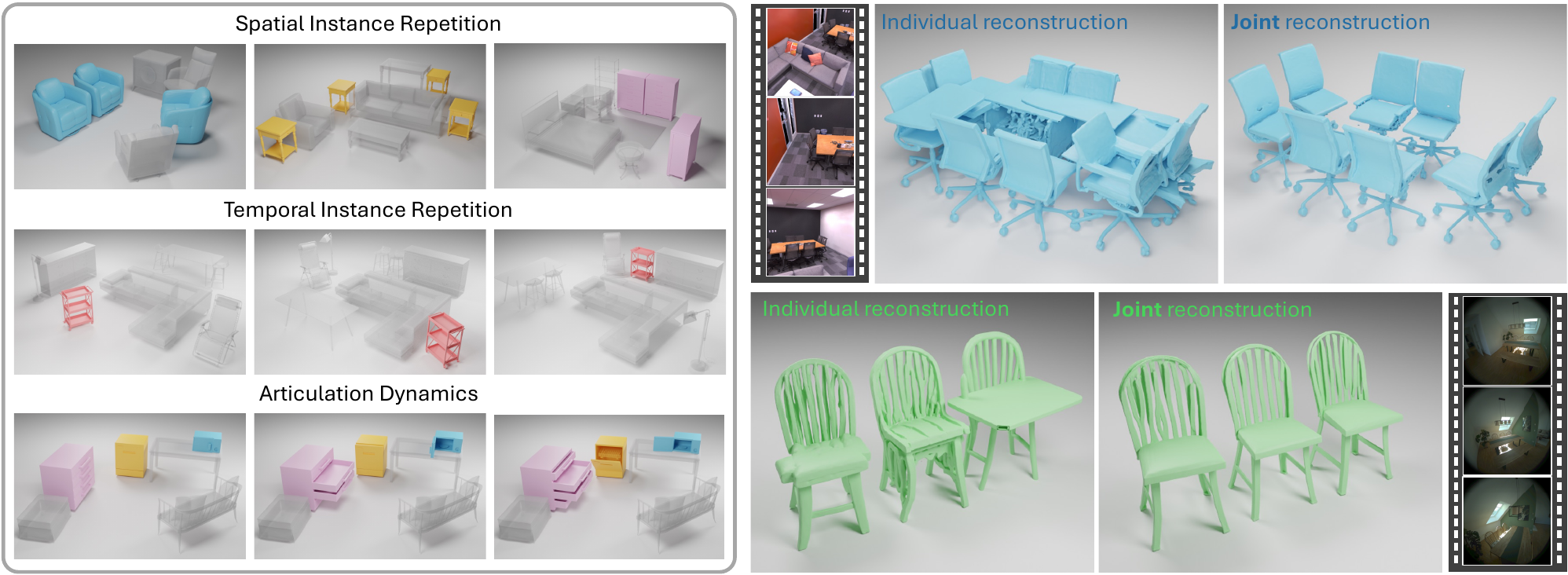}
\captionof{figure}{
We address the challenge of compositional scene reconstruction with objects re-observed across space and time.
We characterize this into three concrete cases (\emph{left}): spatial repetition, temporal repetition and articulation dynamics. 
We propose the Joint Reconstruction Model (\ours) to perform coupled reconstruction of a group of objects, out-performing reconstruction of each individually (\emph{right}). 
}
\label{fig:teaser}
\end{center}
}

\twocolumn[{
\vspace*{-0.5cm}
\maketitle

\vspace{-2.em}
\begin{center}
\captionsetup{type=figure}
\includegraphics[width=\textwidth]{fig/images/teaser.pdf}
\captionof{figure}{
We address the challenge of compositional scene reconstruction with objects re-observed across space and time.
We characterize this into three concrete cases (\emph{left}): spatial repetition, temporal repetition and articulation dynamics. 
We propose the Joint Reconstruction Model (\ours) to perform coupled reconstruction of a group of objects, out-performing reconstruction of each individually (\emph{right}). 
}
\label{fig:teaser}
\end{center}

}]

\def\thefootnote{\fnsymbol{footnote}}
\footnotetext[2]{Work done during internship at Meta.} 
\footnotetext[0]{Project page: \url{https://qiruiw.github.io/jrm}}
\def\thefootnote{\arabic{footnote}}

\begin{abstract}
Object-centric reconstruction seeks to recover the 3D structure of a scene through composition of independent objects. While this independence can simplify modeling, it discards strong signals that could improve reconstruction, notably \textit{repetition} where the same object model is seen multiple times in a scene, or across scans.
We propose the Joint Reconstruction Model (JRM) to leverage repetition by framing object reconstruction as one of personalized generation: multiple observations share a common subject that should be consistent for all observations, while still adhering to the specific pose and state from each.
Prior methods in this direction rely on explicit matching and rigid alignment across observations, making them sensitive to errors and difficult to extend to non-rigid transformations. In contrast, JRM is a 3D flow-matching generative model that implicitly aggregates unaligned observations in its latent space, learning to produce consistent and faithful reconstructions in a data-driven manner without explicit constraints.
Evaluations on synthetic and real-world data show that JRM’s implicit aggregation removes the need for explicit alignment, improves robustness to incorrect associations, and naturally handles non-rigid changes such as articulation. Overall, JRM outperforms both independent and alignment-based baselines in reconstruction quality.
\end{abstract}
\section{Introduction}
\label{sec:intro}


Reconstruction of a 3D scene from visual input provides a foundation for understanding and embodied interaction within the physical space. An object-centric approach, which represents a scene as a collection of individually complete objects, is especially valuable. This representation naturally supports object-level interaction and editing, as well as simplifies modeling by allowing each object to be reconstructed independently.

Recent advances in 3D generative modeling for object reconstruction mean that one can now reasonably expect a faithful and high-fidelity reconstruction from a clear, unoccluded image(s) of an object from a number of different methods~\cite{hunyuan3d22025tencent,siddiqui2024meta,li2025triposg,xiang2024structured}. When paired with additional input modalities, as well as targeted training and augmentation strategies, these same generative architectures maintain fidelity while becoming significantly more robust to the difficult setting of casual egocentric inputs where observations are frequently incomplete or otherwise challenging~\cite{siddiqui2026shaper}.

However, independent reconstruction of each object imposes an important limitation. In real-world scenes, objects rarely appear in isolation. Independent object-centric reconstruction discards rich contextual cues provided by surrounding objects, the broader scene, and previous observations, leading to information loss, especially in complex or cluttered environments.

There are countless examples of such contextual cues from surrounding objects that can enhance reconstruction, such as physical plausibility due to support and intersection. 
We isolate the specific cases of spatial and temporal instance repetition (illustrated in ~\cref{fig:teaser}).
For spatial repetition, consider chairs around a dining table; unless thoroughly scanned, occlusions mean that each is likely observed only partially, in turn hampering independent reconstruction. However, if we can leverage knowledge that the chairs are the same model, the reconstruction can be conditioned on the aggregate of their observations.
Analogously, temporal repetition occurs when multiple sparse scans of a changing scene lead to re-observation of the same object, even if it has moved or deformed. Integration of observations across scans again has the potential to improve the reconstruction in each.


Existing approaches have attempted to address this by explicitly matching, aligning, and registering object instances across scans to integrate observations~\cite{zhu2023living}. However, such explicit methods are sensitive to errors at each stage and struggle to handle sub-object transformations, such as a drawer being opened between scans.

Inspired by recent advances in personalized image generation~\cite{zeng2024jedi}, where multiple consistent images are generated of a common subject, we frame the target object as a shared subject across multiple observations that must be consistently reconstructed in each. To do so, we propose a novel framework that implicitly aggregates unaligned object observations within the latent space of a flow-matching 3D generative model, enabling joint reconstruction of multiple related objects, where the appropriate aggregation across matched instances is learned in a data-driven fashion.

Across both synthetic and real-world datasets, we demonstrate improved reconstruction for both repeated static objects and articulated objects across different states. In addition, we show this learned aggregation is more robust under non-identical object correspondences compared to explicit alignment and registration.

\begin{figure}[t]
\centering
\includegraphics[width=\linewidth]{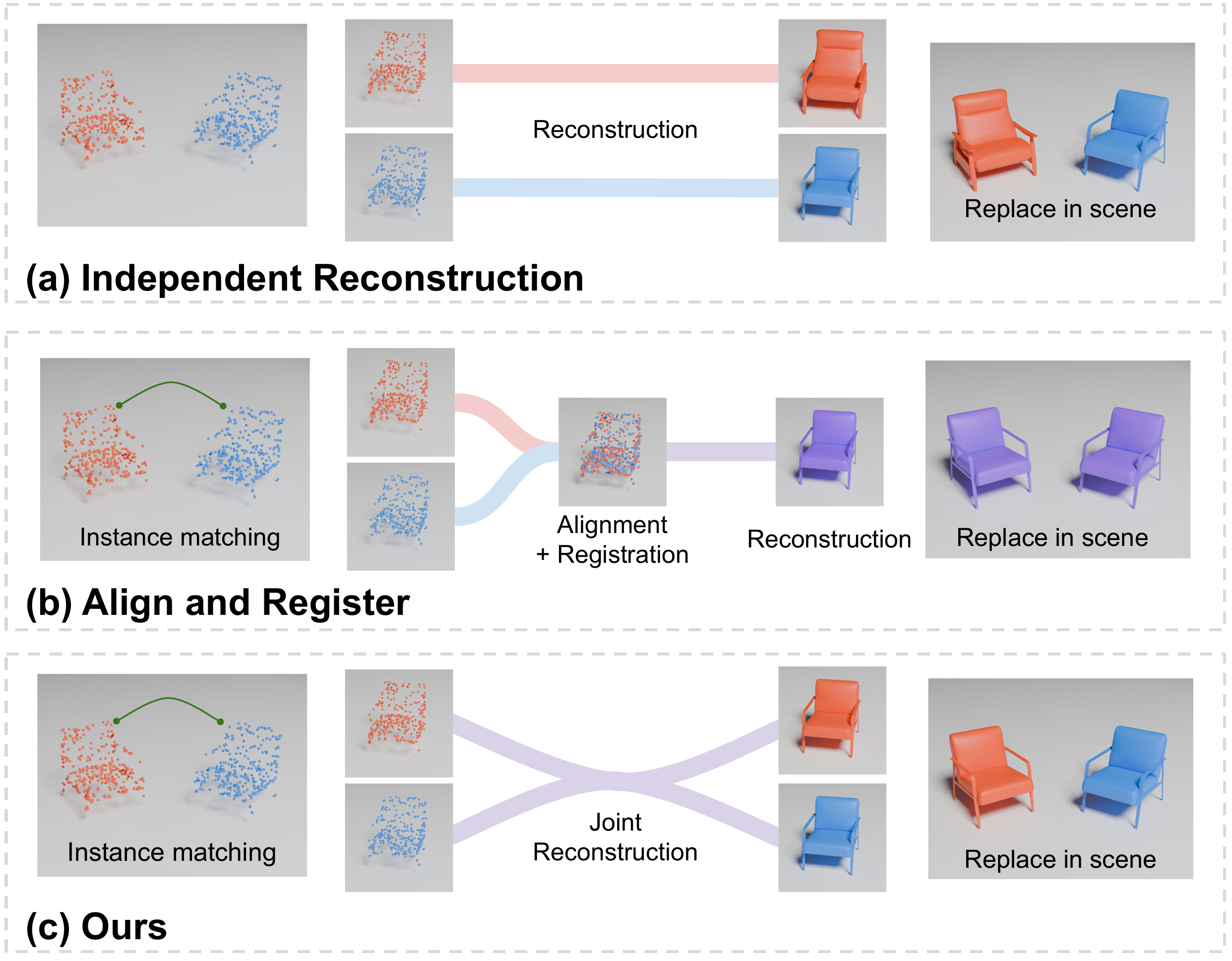}
\caption{
Comparison between different approaches to object-centric reconstruction. \ours offers a relaxation of explicit alignment and registration techniques. Objects are jointly reconstructed, allowing information flow between them, but without imposing hard constraints on similarity.} 
\vspace{-8pt}
\label{fig:pipeline}
\end{figure}

In summary, our contributions are:
\begin{itemize}
    \item We propose JRM, a model that jointly reconstructs a set of unaligned objects through an implicit coupling in the generation latent space. Our approach provides a general framework for repetition-aware reconstruction.
    \item With targeted experiments, we show that this implicit approach exhibits greater robustness to matching and alignment errors than previous explicit methods, as well as demonstrating its extension to sub-object changes, such as articulation.
    \item Through synthetic and real-world evaluation, we demonstrate that this method outperforms previous strategies for compositional reconstruction.
\end{itemize}
\begin{figure*}[t]
\centering
\includegraphics[width=\linewidth]{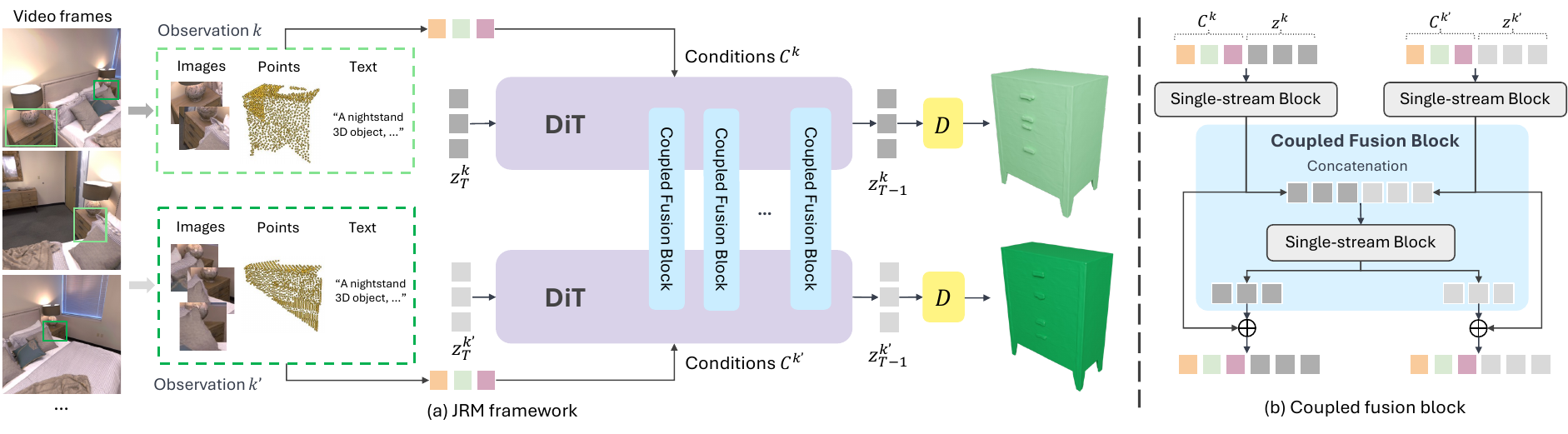}
\caption{(a) \ours jointly reconstructs two nightstands $k$ and $k'$ that appear in a single scan.
(b) Our proposed coupled fusion block, in which the denoised tokens of two distinct objects attend to each other in the latent space to implicitly aggregate unaligned observations. 
} 
\vspace{-8pt}
\label{fig:method}
\end{figure*}

\section{Related Work}

\mypara{Generative priors for compositional reconstruction.}
Early approaches convert a single image to a compositional 3D scene via retrieval-and-alignment~\cite{kuo2020mask2cad, gumeli2022roca, gao2023diffcad, dai2024acdc, wu2024diorama} strategies or pixel-aligned reconstruction~\cite{nie2020total3dunderstanding, zhang2021holistic, liu2022towards, chen2024single} techniques. 
They typically lead to artifacts such as inaccurate or over-smoothed geometry, and unrealistic appearance. These are mainly due to the limited diversity of 3D assets for retrieval, insufficient training data, and the gap between synthetic and realistic domains.

Recent works~\cite{dogaru2024generalizable, zhou2024zero, gu2025artiscene, wang2024architect, yao2025cast} employ pretrained generative models to achieve more thorough and detailed 3D modeling. These largely involve a modular training-free pipeline, where offline 2D image inpainting and 3D mesh generation models are used in sequence for each object, which are prone to the error accumulation. 

Other works~\cite{huang2025midi, zhao2025depr, meng2025scenegen} learn a distribution for occluded objects in cluttered environments by training on full synthetic scenes in an end-to-end manner. However, such reliance restricts their scalability due to the scarcity of comprehensive scene datasets.
Additionally, these methods specifically tailor their design for single-view input, limiting their applicability in multi-view or video contexts.

DPRecon~\cite{ni2025dprecon} accommodates multi-view inputs by employing neural implicit scene representation, similar to prior works~\cite{wu2023objectsdf++, ni2024phyrecon}. It incorporates 2D generative priors to help optimize each object through the SDS loss. While these additional priors improve visually appealing novel view synthesis, the resulting meshes often feature noisy geometry and the additional overhead means the per-scene NeRF-based optimization process takes hours to complete.

In contrast, our method is capable of generating more sharp and geometrically accurate 3D shapes from an unconstrained number of unaligned observations. We further adopt a pair-wise training strategy enabling inter-object patterns to be learned, without requiring full scene supervision.

\mypara{Diffusion models for multiple instances.}
Several studies~\cite{zeng2024jedi, feng2025personalize, kumari2025generating, meng2025scenegen, huang2025midi, yan2025xpart} investigate the mechanism of simultaneously generating multiple instances within the framework of diffusion models for both the 2D and 3D domains. For personalized image generation, JeDi~\cite{zeng2024jedi} generates related images of the same custom subject by coupling tokens across images in a modified self-attention layer of the diffusion U-Net. 
MIDI-3D~\cite{huang2025midi} and SceneGen~\cite{meng2025scenegen} address the challenge of generating multiple 3D objects from a single scene image in one inference. They introduce a multi-instance attention mechanism among all objects that transforms the generation of multiple objects from individual processes into a synchronous interactive process. In the analogous setting of object-centric scene reconstruction, recent works~\cite{yan2025xpart, dong2025copart, ding2025fullpart} on parts-to-object generation also implement both inter- and intra-part level attention layers.

However, these works mainly leverage multi-instance attention to improve spatial arrangement and coherence, but do not clearly show how multi-object generation improves geometric modeling. Furthermore, because scene-level models require complete scenes for training, their scalability is limited by the scarcity of such datasets. JRM's training, conversely, scales with the number of individual assets, enabling training at a much larger scale while still capturing inter-object interactions. By isolating repetition rather than opaque scene-level priors, we systematically demonstrate and analyze the benefits of the joint generation paradigm using carefully targeted synthetic benchmarks.


\mypara{Reconstruction of repeated instances.}
Integrating repeated observations of the same object has largely relied on explicit alignment. LivingScenes~\cite{zhu2023living} proposes a multi-stage pipeline that sequentially solves the sub-problems of object association, point cloud registration, and object reconstruction across separate captures at irregular intervals. Splat-and-Replace~\cite{violante2025splat} improves 3D Gaussian Splatting (3DGS) reconstruction fidelity by realigning repetitive objects from multiple views into a shared coordinate system using estimated rigid transformations for each instance.

This work bypasses the need to explicitly align inputs in the observation space, where mismatched object geometry, texture, and background can cause significant degradation. Instead, we exploit the implicit integration of unaligned observations in the latent space of generative models.

\section{Method}
\newcommand{\latentdim}{L}

We perform \textit{compositional} 3D scene reconstruction, representing the scene as a set of discrete, posed object meshes rather than a single monolithic mesh. To do this, we detect and reconstruct each object from multimodal inputs (observations). We assume camera poses, depths (or point clouds), and instance segmentations are provided by upstream pipelines. As a limitation, reconstruction quality is inherently tied to the accuracy of these provided inputs.

\begin{figure}[t]
\centering
\includegraphics[width=\linewidth]{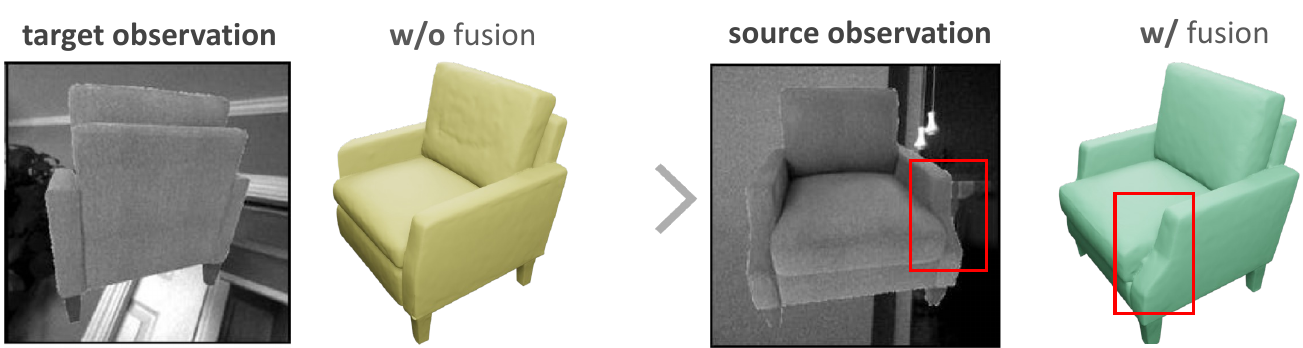}
\caption{
An example \ours reconstruction illustrating the benefit of joint reconstruction. From the first view alone it is not possible to discern the shape of the armrest. However, by adding in a source observation that observes the front, \ours is able to reconstruct the target view correctly. 
} 
\label{fig:fusion}
\end{figure}

\ours extends ShapeR~\cite{siddiqui2026shaper}, a conditional 3D generative model that reconstructs objects individually, by introducing coupled object reconstruction. Jointly reconstructing multiple objects enables information flow between them to improve quality (an illustrative example is shown in~\cref{fig:fusion}).

In this section, we first summarize the ShapeR~\cite{siddiqui2026shaper} architecture and inputs. We then explain its extension for joint object reconstruction.
Specific implementation details are included in the supplement.

\subsection{Multimodally Conditioned Flow Matching }
The ShapeR model~\cite{siddiqui2026shaper} reconstructs a single object from egocentric observations. A VecSets-based~\cite{zhang20233dshape2vecset} variational autoencoder is first trained, allowing a 3D mesh $S$ to be embedded as a set of $n$, $\latentdim$-dimensional latent tokens $z \sim q(z|S), \quad z  \in \mathbb{R}^{n\times \latentdim}$. Latents may be decoded by a decoder $D$ by predicting signed-distance values $s=D(z, x)$ for a grid of query points $x \in \mathbb{R}^3$. 

A denoising Diffusion Transformer (DiT) is trained via rectified flow-matching to transport latents drawn from a standard normal distribution $z_1 \sim \mathcal{N}(0,I)$ to the manifold $z_0$ defined by the distribution of embedded training shapes. 

To reconstruct an object of interest, the denoising process is conditioned on a multimodal \textit{observation}, encoded as $C$, comprising the following signals. The object's segmented partial pointcloud, sparse views, and VLM-derived text descriptions are encoded using a light-weight SparseConvNet~\cite{SubmanifoldSparseConvNet} encoder, frozen DinoV2~\cite{oquab2023dinov2}, and pretrained T5~\cite{raffel2020exploring} encoder, respectively.

\subsection{Joint Reconstruction Model}

Our \ours framework extends the DiT modules used in ShapeR in two ways: it simultaneously denoises the latent representations of multiple 3D objects using a shared DiT network; and it incorporates a new coupled fusion block to facilitate inter-instance interaction, allowing the implicit coupling of generated shapes within the high-dimensional diffusion latent space.

\mypara{Joint diffusion architecture.}
Our proposed \ours simultaneously generates latent tokens for all $K$ objects in an object group $\mathcal{Z} = \{z^k \in \mathbb{R}^{n\times \latentdim}\}^K_1$. As depicted in~\cref{fig:method}, latent tokens $z^k$ along with the corresponding observation condition tokens $C^k$ are passed through the DiT module whose weights are shared for each object.


\mypara{Coupled fusion block.} To allow information flow \emph{between} objects within a given group, we introduce a coupled attention mechanism within the single-stream block of the original DiT architecture. For an original DiT model consisting of a number of single-stream blocks, we replace every alternative single-stream block with a coupled fusion block.

The coupled fusion block enables interaction between jointly denoised shape latents through attention. We concatenate the latent tokens of all objects $z_O = \oplus\{z^k\}^K_1$ and process them with a single-stream block, allowing latents from all objects within a group to attend to each other. The latent tokens for each shape are recovered by splitting the combined output along the original concatenated dimension. We highlight the fact that while individual DiT single-stream blocks process both latents and tokenized observations from individual objects, coupled attention is applied solely between the shape latents of the group of objects. This is mainly following intuition that the subject of generation (the shape latent) should be consistent within a group while each specific instance of an object should respect its own individual observations. 

\subsection{Pair-wise Training for Joint Reconstruction}
\label{sec:pairwise}
In a strict object-centric formulation, each object is reconstructed independently, allowing reconstruction methods to be trained on large datasets of 3D assets where each object is observed in isolation. \ours departs from this paradigm, enabling groups of object-observations as input. Rather than extracting sets of objects from full-scene or multi-scan datasets, which are relatively scarce, \ours is trained on pairs of objects, each observed independently. This ability to extend to an arbitrary number of observed objects during inference, despite only training on pairs of objects, is enabled by \ours's attention-based coupling strategy.



Training examples consist of pairs of objects, which may be either similar or different to one another. We compute cosine similarity between object shape embeddings extracted via DuoDuoCLIP~\cite{lee2025duoduo} from 12 random rendered views. Pairs of objects are considered similar (or positive) if this similarity exceeds $0.9$, otherwise they are classed as different (or negative). During training, we sample negative pairs with a probability of $0.1$ to encourage the model to learn to adaptively aggregate relevant information from the other branch.

For each object in the pair, we use the standard flow matching objective~\cite{lipman2022flow, esser2024scaling}. \ours is trained to predict the velocity field $v_t^k = \frac{dz_t^k}{dt}$ that moves a sample $z_t^k$ towards data $z_0^k$. Concretely, we adopt the affine path with the conditional optimal transport schedule such that $z_t^k = (1- t)\cdot z_0^k + t\cdot \varepsilon^k \text{ and } v_t^k = z_0^k- \varepsilon^k$ according to a noise level $t$ from 0 to 1 and multi-modal conditions $C^k$:
\begin{equation}
    L(\theta) = E_{z^k,C^k,\varepsilon^k}\left[ \sum_{k=1}^2||v_t^k - v_{\theta}(z_t^k, t, C^k)||^2 \right]
\end{equation}

We provide further details of preparation of multimodal conditions for each object in the supplement.

\section{Experiments}

\begin{table*}[t]
\centering
\caption{Quantitative results of temporal instance repetition with ground truth object matching. Numbers in {\color{lightgray}gray color} represent results of applying the oracle object alignment to the condition inputs of \fm.}
\label{tab:livingscenes-gtm}
\small
    \begin{tabular}{@{}l ccc ccc ccc ccc @{}}
        \toprule
        \multirow{2}{*}{Methods} & \multicolumn{3}{c}{Modality} & \multicolumn{3}{c}{Target-only} & \multicolumn{3}{c}{1 Rescan} & \multicolumn{3}{c}{3 Rescans}\\
        \cmidrule(lr){2-4} \cmidrule(lr){5-7} \cmidrule(lr){8-10} \cmidrule(lr){11-13}
        & Point & Image & Text & CD$\downarrow$ & NC$\uparrow$ & F1$\uparrow$ & CD$\downarrow$ & NC$\uparrow$ & F1$\uparrow$ & CD$\downarrow$ & NC$\uparrow$ & F1$\uparrow$ \\
        \midrule
        MORE$^2$ & \cmark & \xmark & \xmark & 10.43 & 74.45 & 32.25 & 9.89 & 74.28 & 33.06 & 10.04 & 73.70 & 34.17 \\
        \multirow{2}{*}{\fm} & \multirow{2}{*}{\cmark} & \multirow{2}{*}{\xmark} & \multirow{2}{*}{\xmark} & \multirow{2}{*}{\textbf{3.07}} & \multirow{2}{*}{\textbf{83.42}} & \multirow{2}{*}{\textbf{86.10}} & 3.71 & 79.16 & 80.50 & 4.43 & 73.63 & 73.68 \\
         & & & & & & & {\color{lightgray}2.35} & {\color{lightgray}85.07} & {\color{lightgray}90.77} & {\color{lightgray}1.98} & {\color{lightgray}85.98} & {\color{lightgray}93.56} \\
        \ours & \cmark & \xmark & \xmark & 3.46 & 83.26 & 85.20 & \textbf{2.95} & \textbf{84.88} & \textbf{87.83} & \textbf{3.07} & \textbf{84.80} & \textbf{86.91} \\
        \midrule
        \multirow{2}{*}{\fm} & \multirow{2}{*}{\cmark} & \multirow{2}{*}{\cmark} & \multirow{2}{*}{\cmark} & \multirow{2}{*}{3.12} & \multirow{2}{*}{\textbf{84.19}} & \multirow{2}{*}{\textbf{88.57}} & 3.50 & 81.92 & 84.55 & 3.62 & 80.09 & 81.17 \\
         & & & & & & & {\color{lightgray}2.52} & {\color{lightgray}85.99} & {\color{lightgray}92.40} & {\color{lightgray}2.03} & {\color{lightgray}87.18} & {\color{lightgray}94.98} \\
        \ours & \cmark & \cmark & \cmark & \textbf{2.84} & 81.75 & 86.74 & \textbf{2.55} & \textbf{83.58} & \textbf{89.11} & \textbf{2.49} & \textbf{83.80} & \textbf{89.24} \\
        \bottomrule
    \end{tabular}
\end{table*}
\begin{table*}[!t]
\caption{Quantitative results of spatial instance repetition where target objects are matched with different types of source objects. Numbers in {\color{lightgray}gray color} indicate applying the oracle alignment to the observations for \fm.}
\label{tab:controlscenes}
\centering
\small
    \begin{tabular}{@{}l ccc ccc ccc ccc @{}}
    \toprule
    \multirow{2}{*}{Methods} & \multicolumn{3}{c}{Target-only} & \multicolumn{3}{c}{Identical Pair} & \multicolumn{3}{c}{Similar Pair} & \multicolumn{3}{c}{Negative Pair}\\
    \cmidrule(lr){2-4} \cmidrule(lr){5-7} \cmidrule(lr){8-10} \cmidrule(lr){11-13}
    & CD$\downarrow$ & NC$\uparrow$ & F1$\uparrow$ & CD$\downarrow$ & NC$\uparrow$ & F1$\uparrow$ & CD$\downarrow$ & NC$\uparrow$ & F1$\uparrow$ & CD$\downarrow$ & NC$\uparrow$ & F1$\uparrow$ \\
    \midrule
    \fm & 3.33 & \textbf{83.83} & 87.04 & {\color{lightgray}2.77} & {\color{lightgray}85.46} & {\color{lightgray}90.91} & {\color{lightgray}4.78} & {\color{lightgray}77.11} & {\color{lightgray}72.51} & {\color{lightgray}8.83} & {\color{lightgray}72.99} & {\color{lightgray}59.22} \\
    \ours & \textbf{2.61} & 82.26 & \textbf{87.99} & \textbf{2.49} & \textbf{82.93} & \textbf{88.70} & \textbf{2.72} & \textbf{82.09} & \textbf{86.90} & \textbf{3.04} & \textbf{80.89} & \textbf{86.31} \\
    \bottomrule
    \end{tabular}
\end{table*}
\begin{figure*}
\centering
\includegraphics[width=\linewidth,trim={0 4px 0 5px},clip]{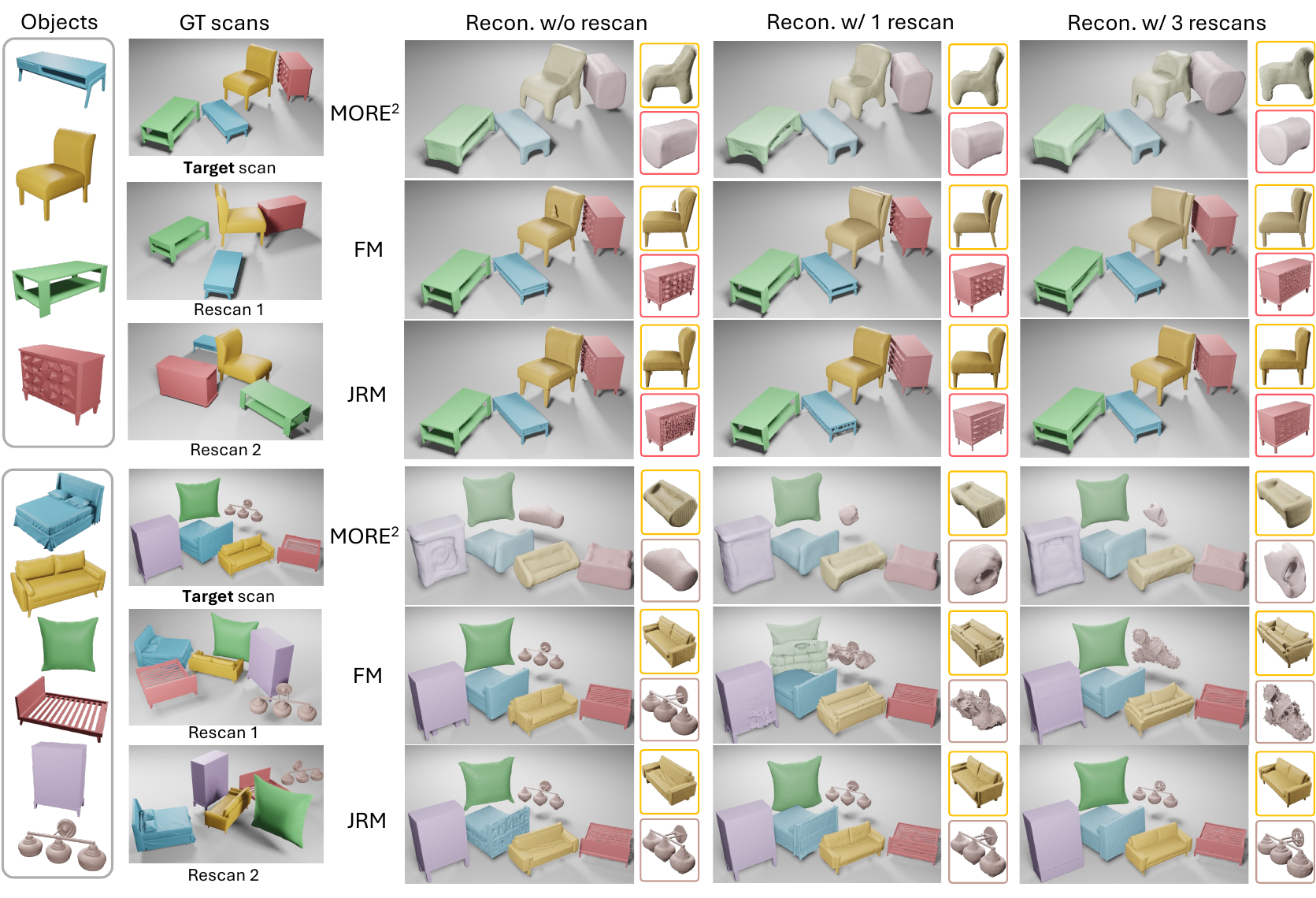}
\caption{
Qualitative results on temporal instance repetition. Objects are colored from {\color{lightgray} \emph{faded}} to \textbf{\emph{solid}} tones reflecting chamfer distance to the ground truth. The reconstruction from \ours improves with more rescans, while those from \fm degrade due to inaccurate alignments. 
} 
\vspace{-8pt}
\label{fig:livingscenes}
\end{figure*}


We evaluate \ours for its ability to reconstruct objects in various instance repetition scenarios. In all experiments, we compute reconstruction metrics on a set of target objects. In the individual reconstruction scenario, target objects are reconstructed conditioned solely on the multimodal observations of the target itself. To evaluate the case of joint reconstruction, we introduce additional support sets of object observations to reconstruct jointly with the target.\footnote{Although \ours also produces reconstructions for these support objects, we still evaluate only the reconstruction of the target objects.} The formation of this support set is experiment-dependent, with details provided in each respective section.

\mypara{Metrics.}
We compute the Chamfer Distance (CD, in cm), Normal Consistency (NC), and F-Score following~\cite{zhu2023living, ni2025dprecon}.

\mypara{Baselines.}
Our main comparison is a smaller variant of ShapeR~\cite{siddiqui2026shaper}.
To ensure a fair comparison, we train both \fm and \ours on the identical set of objects and ensure that the models share approximately the same capacity, each consisting of 24 transformer blocks in total. For the \fm baseline, observations of support objects are rigidly aligned with the target's based on relative pose and concatenated.
We also compare with a pre-trained version of MORE$^2$~\cite{zhu2023living} on temporal rescan inputs and with DPRecon~\cite{ni2025dprecon} on real-world scenes.

\mypara{Datasets.}
The initial object-centric training of the base \fm architecture is conducted on a pool of 400k high-quality 3D objects from a data mix that combines ObjaverseXL~\cite{deitke2023objaverse}, Amazon Berkeley Objects~\cite{collins2022abo}, Wayfair~\cite{wayfair2016models} and additional artist-created meshes. An 80k subset of this asset pool is used for the pair-wise training as described in~\cref{sec:pairwise}.

A held-out subset of 2,500 objects are used for synthetic evaluation. These objects are used to construct synthetic scenes through iterative placement with simple heuristics designed to approximate real-world occlusion and clutter. We adopt different composition strategies depending on experiment; each explained in respective sections. 

For real-world evaluation, we use the Replica~\cite{straub2019replica} and ScanNet++~\cite{yeshwanth2023scannet++} following DPRecon~\cite{ni2025dprecon}.

\subsection{Results on Synthetic Scenes}
\label{exp:synthetic}
In this section, we conduct controlled experiments with arrangements of synthetic objects. We first investigate the setting of \emph{temporal repetition} of instances, adopting the setting from LivingScenes~\cite{zhu2023living} where we seek to perform instance-level reconstruction of a changing 3D scene with access to multiple sparse scans through time. Next, we investigate \emph{spatial repetition} of instances, where we leverage repetition of object instances within a single scan. Finally, we investigate the case of objects that have undergone sub-object-level changes via reconstruction of articulating objects.

\subsubsection{Temporal Instance Repetition}

This experiment measures the ability to integrate observations across rescans of the same object.
We generate 100 synthetic scenes with a varied number of objects ranging from 4 to 8. We prepare 1 target scan and 3 additional rescans, with random rigid transforms applied to each object, as sources for reconstruction of the target. 


For all methods, we use an oracle object matcher controlling for the effect of incorrect instance association. For MORE$^2$ and \fm, which each require aligned association, we evaluate using the alignment predicted by MORE$^2$.

\mypara{Results.} The results of temporal aggregation are listed in~\cref{tab:livingscenes-gtm}. It is worth first noting that, with access to the same alignment, the generative \fm outperforms MORE$^2$'s reconstruction. This improvement is attributable to architecture, rather than aggregation strategy.

When reconstructing with the target scan alone, \ours performs comparably to \fm. With further rescans as shown in~\cref{fig:livingscenes}, \ours demonstrates consistent improvement with access to more context. On the other hand, it becomes evident that \fm encounters difficulties in enhancing the reconstruction due to its sensitivity to errors in object alignment.
The gray numbers in~\cref{tab:livingscenes-gtm} reveal the ideal aggregation of observations in the context of conditional generation when an oracle object aligner is used, and confirm this sensitivity.
We investigate the sensitivity of \fm to the alignment of instances between scans in (\cref{fig:align-error}, \textit{left}).




\subsubsection{Spatial Instance Repetition}
To evaluate spatial instance repetition, where a scene consists of multiple, repeated objects, we generate 100 synthetic scenes containing 6 objects each: a target, two occluders, and three source objects (one identical, one similar, and one negative match to the target).
Similar matches are sampled by semantic category and a similarity metric above a threshold. Negative matches are sampled from different semantic categories. Reconstruction metrics are computed on the target object in multiple configurations defined by which, if any, source object is used as additional context for reconstruction.
For the baseline \fm, we use an oracle object aligner.

\mypara{Results.}
Metrics are reported in~\cref{tab:controlscenes}. The quality achieved for independent reconstruction, ``Target-only'', serves as the baseline performance reflecting how different methods can improve with additional observations.

\begin{figure}[t]
\centering
\includegraphics[width=\linewidth]{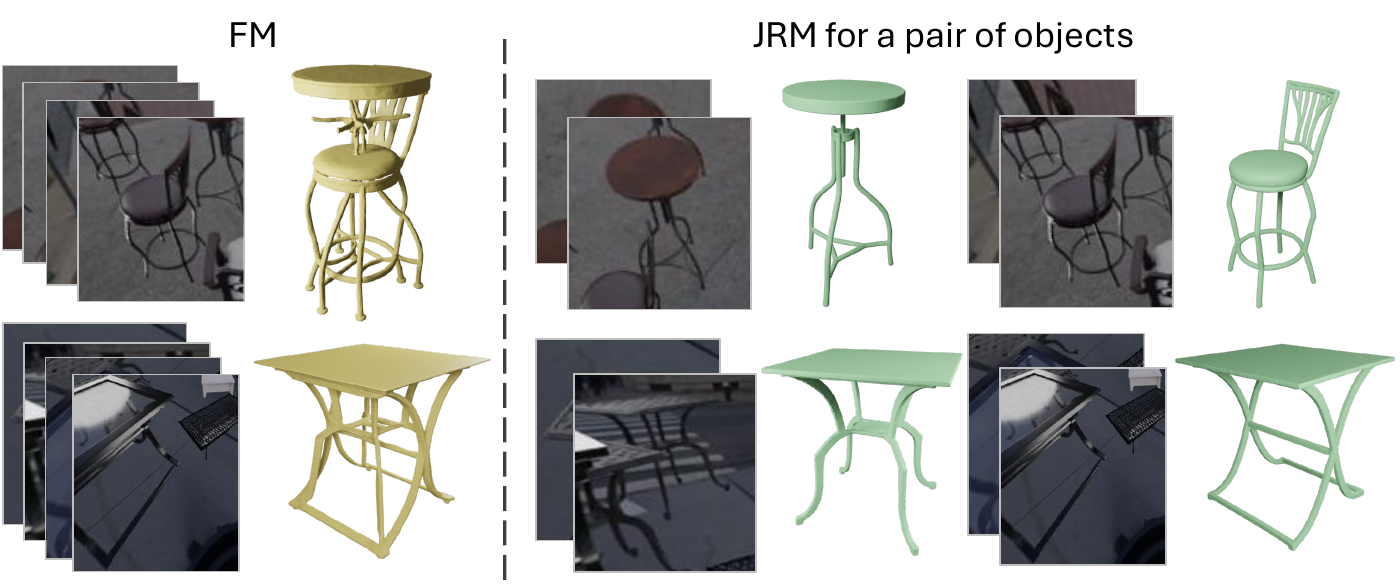}
\caption{
Results of \fm and \ours with mismatched objects. 
\fm naively fuses available observations while \ours presents robustness to incorrect object association.
} 
\vspace{-6pt}
\label{fig:control}
\end{figure}

With an identical object as the source, both \fm and \ours improve, confirming our core motivation that additional information from repetition can improve reconstruction. However, with \fm we see that as we use similar or different objects as sources, we see increasingly bad reconstruction. While not surprising, it does highlight the sensitivity of the chained pipeline to matching errors. Although reconstructions do deteriorate compared to independent reconstruction, \ours demonstrates improved robustness to mismatched, distractor objects. \Cref{fig:control} illustrates the respective reconstructions in a mismatched scenario. 

\begin{figure}[t]
\centering
\includegraphics[width=\linewidth]{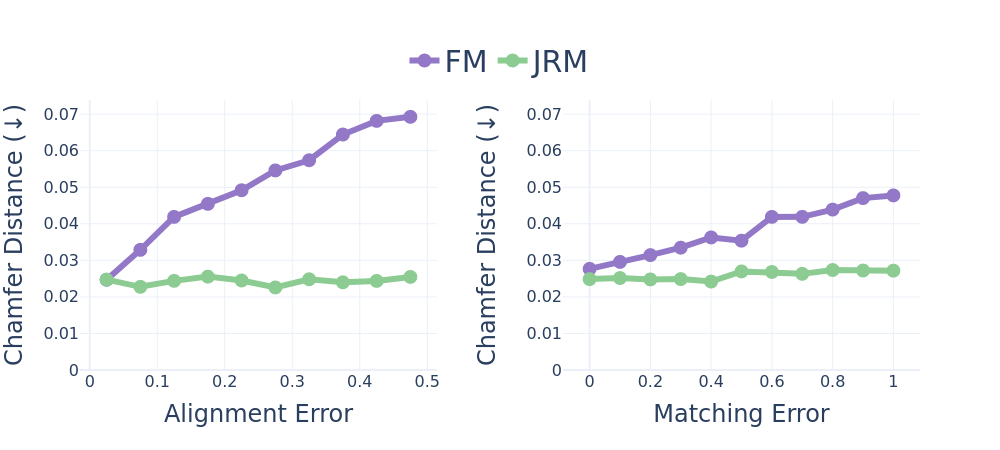}
\caption{
Plot of reconstruction quality versus: alignment error \textit{(left)}; matching error \textit{(right)}. \fm is sensitive to errors in both alignment and matching, while \ours is robust to these errors.
} 
\vspace{-8pt}
\label{fig:align-error}
\end{figure}

We further investigate how the reconstruction quality evolves with an increasing number of incorrectly matched objects across 100 objects (\cref{fig:align-error}, \textit{right}). In this plot we see a stark difference in the sensitivity to matching errors.


\subsubsection{Articulation}
Moving beyond rigid transformations, we now evaluate sub-object level changes by reconstructing articulable objects (e.g., cabinets) across different motion states.
We create 100 scenes, each featuring 3 copies of the same articulated object along with 2 random objects serving as occluders. The 3 copies display one resting state and two randomly chosen articulated states. We fine tune both \fm and \ours using 50k procedurally-generated articulated objects.
Reconstruction metrics are computed for each articulated copy with respect to its own geometry. The \fm baseline makes use of the ground-truth object-level rigid alignment between the 3 articulating objects. For \ours, we jointly reconstruct 3 instances of the articulated object at once.

\begin{figure}[t]
\centering
\includegraphics[width=\linewidth]{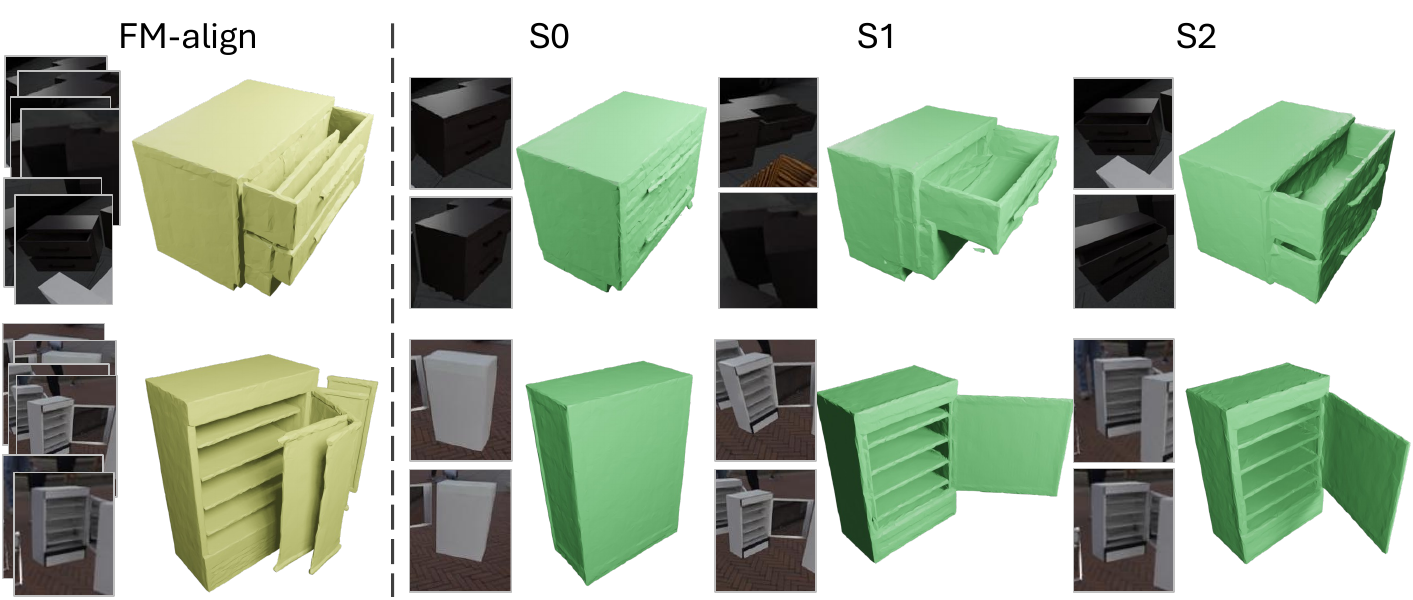}
\caption{
Reconstructions from \fm-align and \ours for a repeated object in 3 different articulation states. 
} 
\vspace{-8pt}
\label{fig:articulated}
\end{figure}

\mypara{Results.}
We report metrics in \cref{tab:artiscenes}. The relatively poor reconstructions from \fm-align emphasize that object-level alignment does not extend well to deforming objects, further illustrated in~\cref{fig:articulated}. On the other hand, \ours achieves noticeably better results on CD and F1 score, while maintaining competitive results on NC. Compared to \fm reconstruction of each object instance independently (\fm-ind.), it is apparent that \ours is able reconstruct consistent, but not necessarily identical, objects.

\subsection{Results on Real-world Scenes}
\label{exp:real}

We evaluate \ours's performance on real-world captures with 7 scenes from Replica~\cite{straub2019replica} and 6 scenes from ScanNet++~\cite{yeshwanth2023scannet++}, where there are repeated objects in environments such as a meeting room. 
Following DPRecon~\cite{ni2025dprecon}, we reconstruct compositional 3D scenes from 10 viewpoints. We use the ground-truth depth for DPRecon per-scene optimization. We obtain the 3D bounding box from the ground-truth mesh segmentation and collect image crops from each for all objects. Objects are grouped by their category. A maximum of 9 objects are reconstructed jointly by \ours, splitting larger groups as necessary.

\begin{figure*}
\centering
\includegraphics[width=\linewidth]{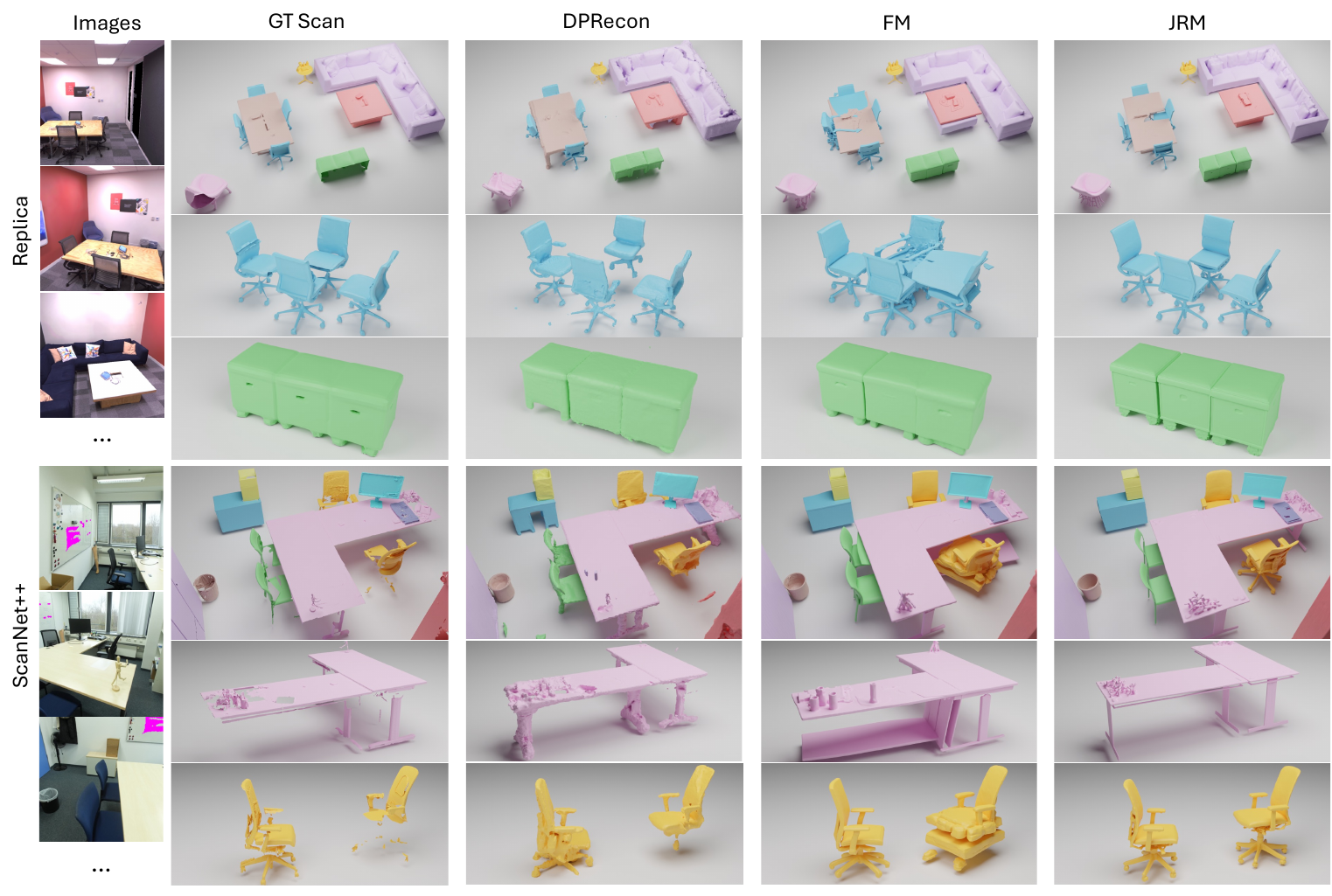}
\caption{
Qualitative results on real-world scenes, including ScanNet++ and Replica.
We use the same color for objects within the same match group. For each scene, we produce two close-up views focusing on the matched objects.
} 
\vspace{-8pt}
\label{fig:realscenes}
\end{figure*}

\mypara{Results.} Reconstructions are depicted in~\cref{fig:realscenes} with metrics compiled in~\cref{tab:real}. We find that \ours's performance holds for real inputs, aggregating information across matched objects to achieve a more consistent and complete geometry. \fm suffers from poor hallucination in occluded areas of objects. DPRecon produces incomplete meshes from its object-wise optimization. Overall, we find that the performance difference between \fm and \ours is larger in ScanNet++ than in Replica. We suspect this is because ScanNet++ test scenes feature more repeated objects.


\begin{table}
\centering
\caption{Quantitative results for repeated articulating objects in different motion states. 
Numbers in {\color{lightgray}gray} reveal results with oracle rigid alignment between objects in different articulation states.}
\label{tab:artiscenes}
\resizebox{\linewidth}{!}
{
\begin{tabular}{@{}l ccc ccc ccc @{}}
\toprule
\multirow{2}{*}{Methods} & \multicolumn{3}{c}{State 0 - Rest} & \multicolumn{3}{c}{State 1} & \multicolumn{3}{c}{State 2} \\
\cmidrule(lr){2-4} \cmidrule(lr){5-7} \cmidrule(lr){8-10} 
& CD$\downarrow$ & NC$\uparrow$ & F1$\uparrow$ & CD$\downarrow$ & NC$\uparrow$ & F1$\uparrow$ & CD$\downarrow$ & NC$\uparrow$ & F1$\uparrow$ \\
\midrule
\fm-align & {\color{lightgray}5.83} & {\color{lightgray}80.15} & {\color{lightgray}70.98} & {\color{lightgray}5.37} & {\color{lightgray}65.68} & {\color{lightgray}76.32} & {\color{lightgray}5.49} & {\color{lightgray}64.91} & {\color{lightgray}76.19} \\
\fm-ind. & 4.92 & \textbf{79.10} & 70.30 & 5.44 & \textbf{77.66} & 76.63 & 4.92 & \textbf{77.98} & 77.42 \\
\ours & \textbf{4.69} & 79.00 & \textbf{73.09} & \textbf{3.66} & 77.51 & \textbf{81.08} & \textbf{3.66} & 77.06 & \textbf{80.81} \\
\bottomrule
\end{tabular}
}
\end{table}
\begin{table}
\caption{Reconstruction results on real scenes, Replica and ScanNet++. 
\ours achieves overall the best performance by only training on pairs of synthetic objects.
}
\label{tab:real}
\centering
\resizebox{0.96\linewidth}{!}
{
\begin{tabular}{@{}l ccc ccc @{}}
\toprule
\multirow{2}{*}{Methods} & \multicolumn{3}{c}{Replica} & \multicolumn{3}{c}{ScanNet++}\\
\cmidrule(lr){2-4} \cmidrule(lr){5-7} 
 & CD$\downarrow$ & NC$\uparrow$ & F1$\uparrow$ & CD$\downarrow$ & NC$\uparrow$ & F1$\uparrow$ \\
\midrule
DPRecon & 4.65 & 74.87 & 71.95 & 5.53 & 72.47 & 65.98 \\
\fm & 3.74 & \textbf{79.28} & 79.21 & 4.20 & 78.60 & 72.96  \\
\ours & \textbf{3.21} & 77.88 & \textbf{81.78} & \textbf{2.69} & \textbf{79.41} & \textbf{85.53} \\
\bottomrule
\end{tabular}
}
\end{table}
\subsection{Training Pair Ablation}
\label{ablation}
We ablate the negative pair training ratio (\cref{tab:neg_ratios}) to evaluate its effect on joint reconstruction. Training solely on similar pairs (ratio 0) makes the model highly sensitive to matching errors. Conversely, training exclusively on negative pairs (ratio 1) causes the model to disregard support objects entirely. Introducing a moderate ratio of negative pairs provides the optimal balance, maintaining similar information aggregation while improving robustness to mismatches.

\begin{table}
\caption{Ablation on negative pair sampling ratio during training.}
\label{tab:neg_ratios}
\centering
\resizebox{\linewidth}{!}
{
\begin{tabular}{@{}l ccc ccc ccc @{}}
\toprule
\multirow{2}{*}{Neg. Ratio} & \multicolumn{3}{c}{Identical Pair} & \multicolumn{3}{c}{Similar Pair} & \multicolumn{3}{c}{Negative Pair}\\
\cmidrule(lr){2-4} \cmidrule(lr){5-7} \cmidrule(lr){8-10}
 & CD$\downarrow$ & NC$\uparrow$ & F1$\uparrow$ & CD$\downarrow$ & NC$\uparrow$ & F1$\uparrow$ & CD$\downarrow$ & NC$\uparrow$ & F1$\uparrow$ \\
\midrule
0.0 & 2.79 & 83.15 & 88.30 & 3.17 & 80.72 & 83.52 & 6.22 & 75.63 & 72.42 \\
0.1 & \textbf{2.18} & \textbf{83.80} & \textbf{91.15} & 2.65 & 82.07 & \textbf{88.34} & \textbf{2.56} & \textbf{82.90} & \textbf{88.69} \\
0.5 & 2.77 & 80.62 & 86.97 & 2.70 & 80.42 & 86.39 & 2.80 & 80.24 & 86.52 \\
0.9 & 2.68 & 80.76 & 86.93 & 2.66 & 81.20 & 87.61 & 2.63 & 80.93 & 87.73 \\
1.0 & 2.69 & 82.21 & 88.07 & \textbf{2.61} & \textbf{82.17} & 87.92 & 2.61 & 82.54 & 88.63 \\
\bottomrule
\end{tabular}
}
\end{table}

\section{Conclusion}

We have presented JRM, a model that jointly reconstructs objects across repeated observations, demonstrating improved reconstruction accuracy on synthetic and real-world benchmarks. In synthetic evaluation, we have shown that JRM's implicit aggregation improves robustness to alignment and association errors in comparison to previous approaches that performed explicit alignment and registration. Although investigated solely in the context of repeated observations, JRM offers a general and flexible reconstruction framework to leverage further cues from the surrounding scene.
{
    \small
    \bibliographystyle{ieeenat_fullname}
    \bibliography{main}
}

\clearpage

\maketitlesupplementary

\appendix

In this supplement, we provide additional details of our method (\Cref{sec:supp-method}) and experiments. We describe the details of synthetic data preparation for training and evaluation benchmarks (\Cref{sec:supp-data}). We present further ablation studies, discuss the benefits of joint reconstruction for better implicit segmentation, and qualitative results of \ours on additional real-world scenes (\Cref{sec:supp-experiments}).

\begin{figure}[t]
\centering
\includegraphics[width=\linewidth]{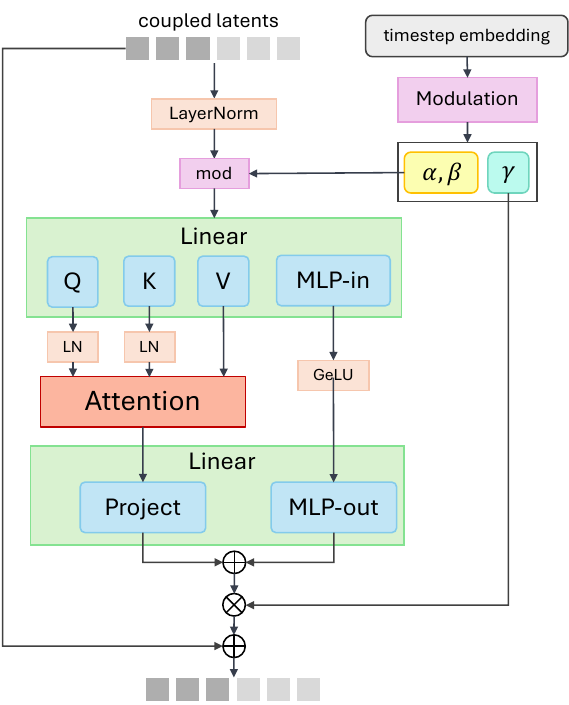}
\caption{
Illustration of a single-stream block taking coupled latents as input.
} 
\label{fig:coupled}
\end{figure}
\begin{figure*}[t]
\centering
\includegraphics[width=\linewidth]{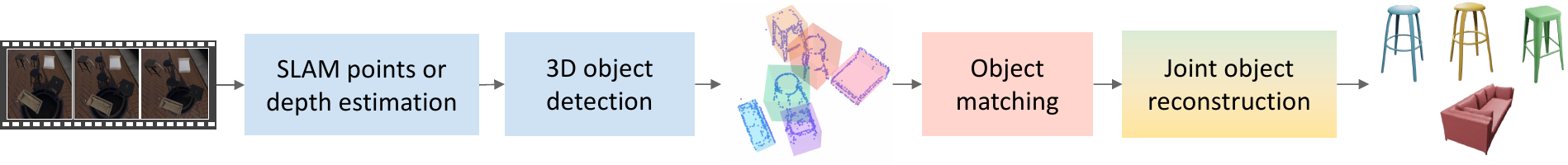}
\caption{
A simple flowchart of the complete pipeline of compositional 3D scene reconstruction from video inputs.
} 
\label{fig:flowchart}
\end{figure*}
\begin{figure}[t]
\centering
\includegraphics[width=\linewidth]{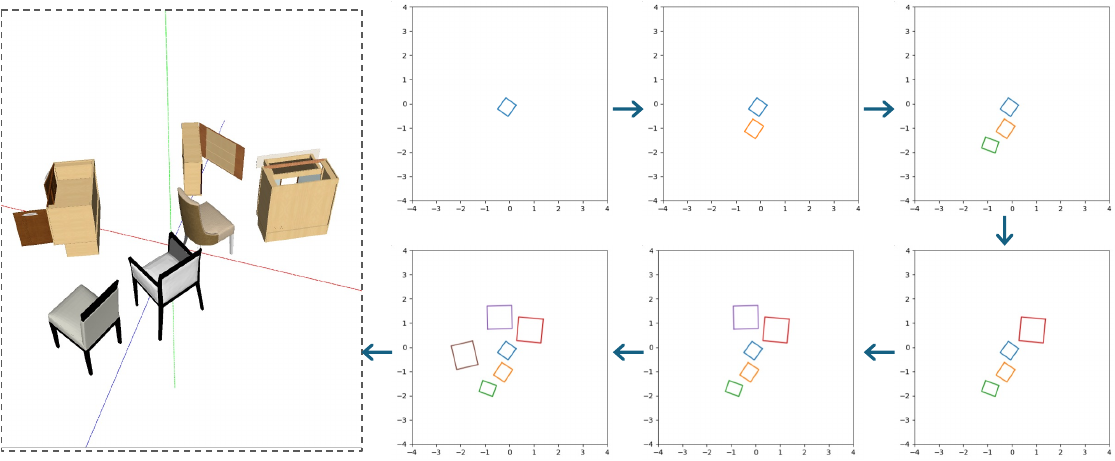}
\caption{
A top-down view illustration of how a synthetic scene is constructed with iterative object insertion by solving collision on 2D polygons (bounding boxes).
} 
\label{fig:heuristics}
\end{figure}
\begin{figure}[t]
\centering
\includegraphics[width=\linewidth]{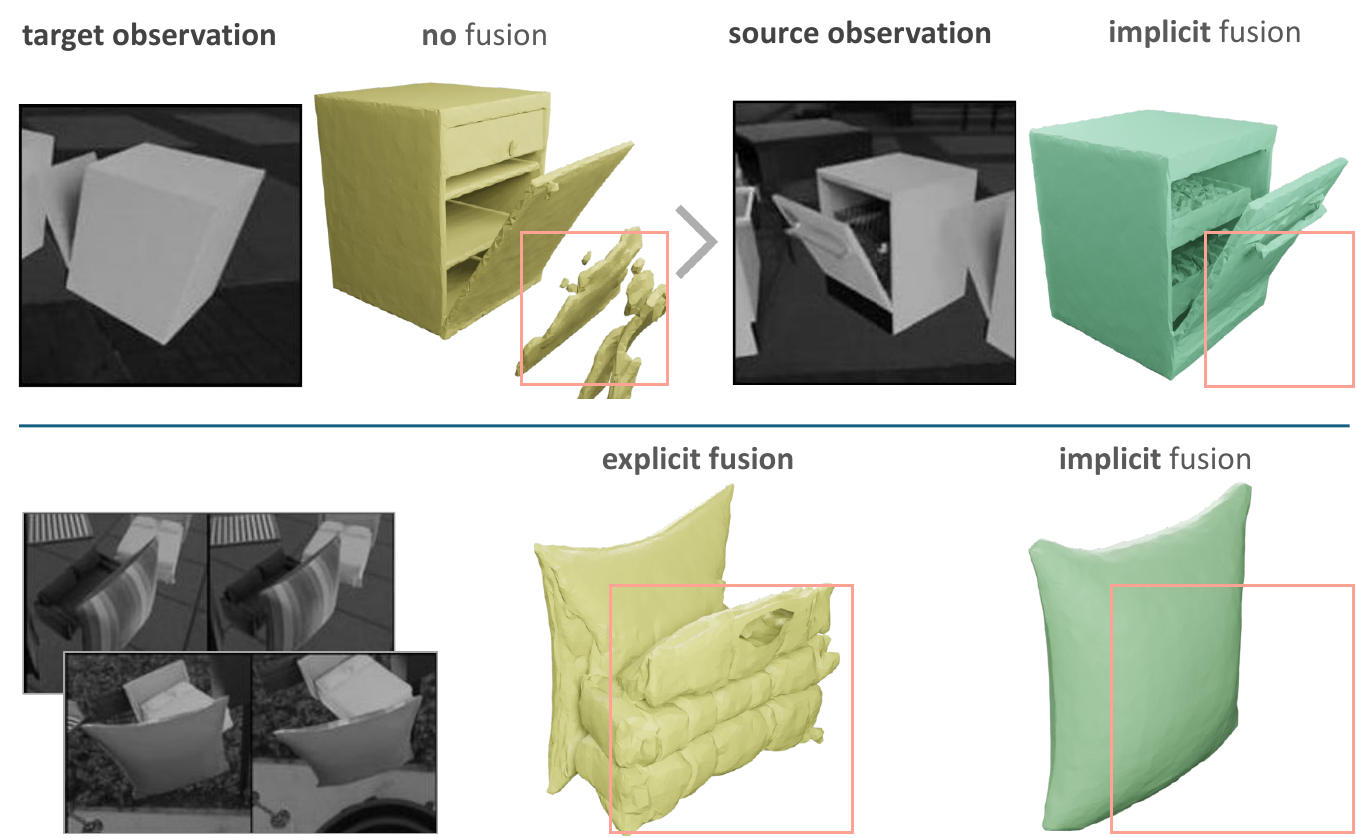}
\caption{
Implicit observation aggregation helps better segmentation during shape generation.
} 
\label{fig:segment}
\end{figure}
\begin{figure*}
\centering
\includegraphics[width=\linewidth]{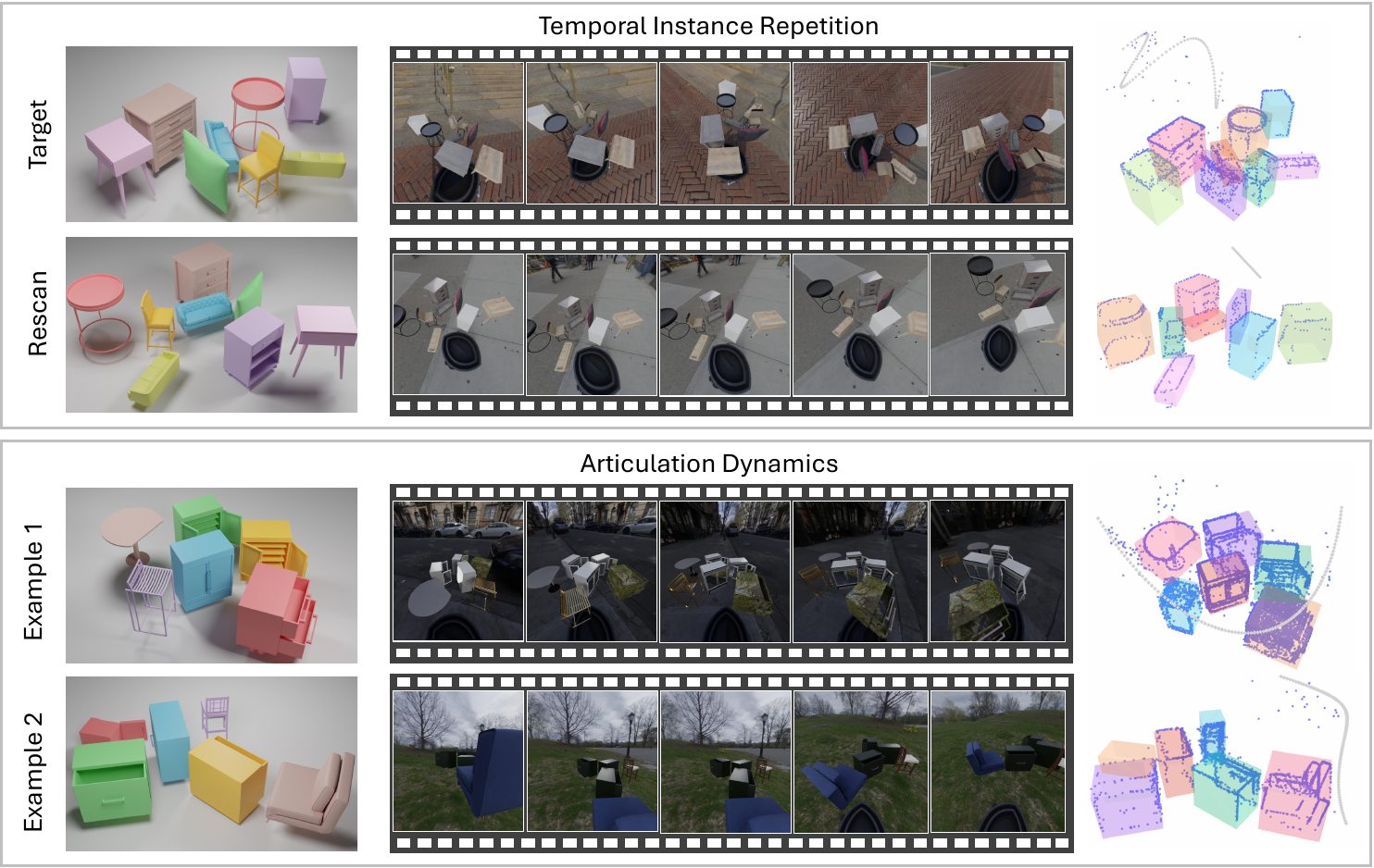}
\caption{
We show examples of generated synthetic scenes in scenarios of temporal instance repetition and articulation dynamics. In the last column, we visualize SLAM points (blue points), camera trajectory (gray diamonds) and 3D bounding boxes.
} 
\label{fig:syn_examples}
\end{figure*}
\begin{table*}
\centering
\caption{Qualitative results on two \ours variants.}
\label{tab:jrm_variants}
\resizebox{\linewidth}{!}
{
\begin{tabular}{@{}l ccc ccc ccc ccc @{}}
\toprule
\multirow{2}{*}{JRM} & \multicolumn{3}{c}{No Pair} & \multicolumn{3}{c}{Identical Pair} & \multicolumn{3}{c}{Similar Pair} & \multicolumn{3}{c}{Negative Pair}\\
\cmidrule(lr){2-4} \cmidrule(lr){5-7} \cmidrule(lr){8-10} \cmidrule(lr){11-13}
& CD$\downarrow$ & NC$\uparrow$ & F1$\uparrow$ & CD$\downarrow$ & NC$\uparrow$ & F1$\uparrow$ & CD$\downarrow$ & NC$\uparrow$ & F1$\uparrow$ & CD$\downarrow$ & NC$\uparrow$ & F1$\uparrow$ \\
\midrule
Replace & 2.61 & 82.26 & 87.99 & 2.49 & 82.93 & 88.70 & 2.72 & 82.09 & 86.90 & 3.04 & 80.89 & 86.31 \\
Insert& \textbf{2.37} & \textbf{82.96} & \textbf{89.62} & \textbf{2.18} & \textbf{83.80} & \textbf{91.15} & \textbf{2.65} & \textbf{82.07} & \textbf{88.34} & \textbf{2.56} & \textbf{82.90} & \textbf{88.69} \\
\bottomrule
\end{tabular}
}
\end{table*}
\begin{table*}
\caption{Quantitative results of temporal instance repetition with \emph{predicted object matching and alignment} by MORE$^2$. Numbers in {\color{lightgray}gray color} represent results of only applying the oracle object alignment to the condition inputs of \fm.
}
\centering
\resizebox{\linewidth}{!}
{
\begin{tabular}{@{}l ccc ccc ccc ccc @{}}
\toprule
\multirow{2}{*}{Methods} & \multicolumn{3}{c}{Modality} & \multicolumn{3}{c}{No Rescan} & \multicolumn{3}{c}{1 Rescan} & \multicolumn{3}{c}{3 Rescans}\\
\cmidrule(lr){2-4} \cmidrule(lr){5-7} \cmidrule(lr){8-10} \cmidrule(lr){11-13}
& Point & Image & Text & CD$\downarrow$ & NC$\uparrow$ & F1$\uparrow$ & CD$\downarrow$ & NC$\uparrow$ & F1$\uparrow$ & CD$\downarrow$ & NC$\uparrow$ & F1$\uparrow$ \\
\midrule
MORE$^2$\cite{zhu2023living} & \cmark & \xmark & \xmark & 10.43 & 74.45 & 32.25 & 10.08 & 74.12 & 32.94 & 10.24 & 73.40 & 33.36 \\
\multirow{2}{*}{\fm} & \multirow{2}{*}{\cmark} & \multirow{2}{*}{\xmark} & \multirow{2}{*}{\xmark} & \multirow{2}{*}{\textbf{3.07}} & \multirow{2}{*}{\textbf{83.42}} & \multirow{2}{*}{\textbf{86.10}} & 3.82 & 78.87 & 80.05 & 4.68 & 72.42 & 71.45 \\
 & & & & & & & {\color{lightgray}2.66} & {\color{lightgray}84.25} & {\color{lightgray}88.86} & {\color{lightgray}2.67} & {\color{lightgray}84.30} & {\color{lightgray}89.38} \\
\ours & \cmark & \xmark & \xmark & 3.46 & 83.26 & 85.20 & \textbf{3.17} & \textbf{84.36} & \textbf{86.35} & \textbf{3.35} & \textbf{84.07} & \textbf{85.57} \\
\midrule
\multirow{2}{*}{\fm} & \multirow{2}{*}{\cmark} & \multirow{2}{*}{\cmark} & \multirow{2}{*}{\cmark} & \multirow{2}{*}{3.12} & \multirow{2}{*}{\textbf{84.19}} & \multirow{2}{*}{\textbf{88.57}} & 3.54 & 81.67 & 84.08 & 3.85 & 79.34 & 79.85 \\
 & & & & & & & {\color{lightgray}2.90} & {\color{lightgray}85.14} & {\color{lightgray}90.18} & {\color{lightgray}2.49} & {\color{lightgray}85.90} & {\color{lightgray}92.01} \\
\ours & \cmark & \cmark & \cmark & \textbf{2.84} & 81.75 & 86.74 & \textbf{2.62} & \textbf{83.17} & \textbf{88.11} & \textbf{2.54} & \textbf{83.57} & \textbf{89.04} \\
\bottomrule
\end{tabular}
}
\label{tab:livingscenes}
\end{table*}
\begin{figure}[t]
\centering
\includegraphics[width=\linewidth]{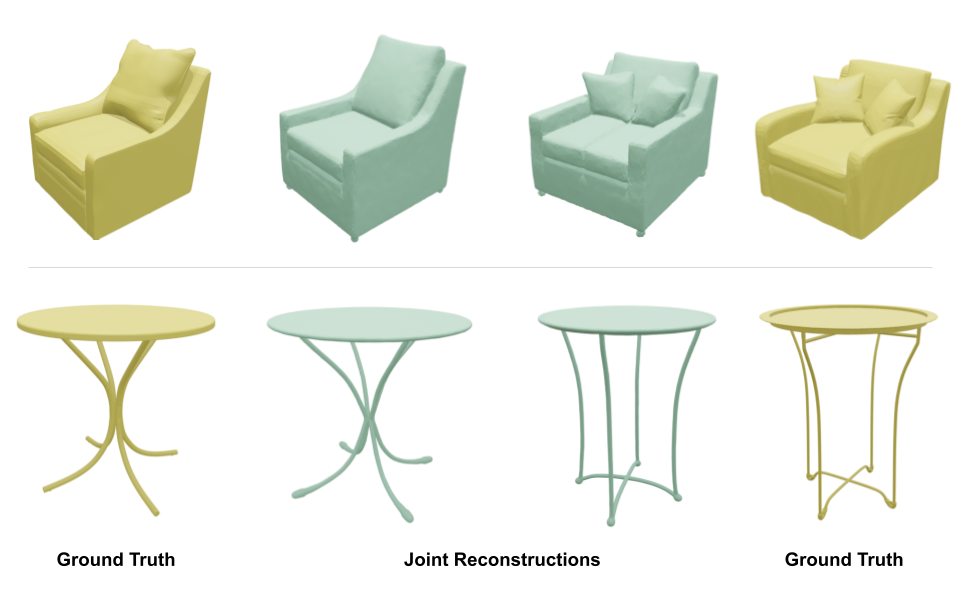}
\caption{
Reconstruction with similar source objects.
} 
\label{fig:similar}
\end{figure}
\begin{figure}[t]
\centering
\includegraphics[width=\linewidth]{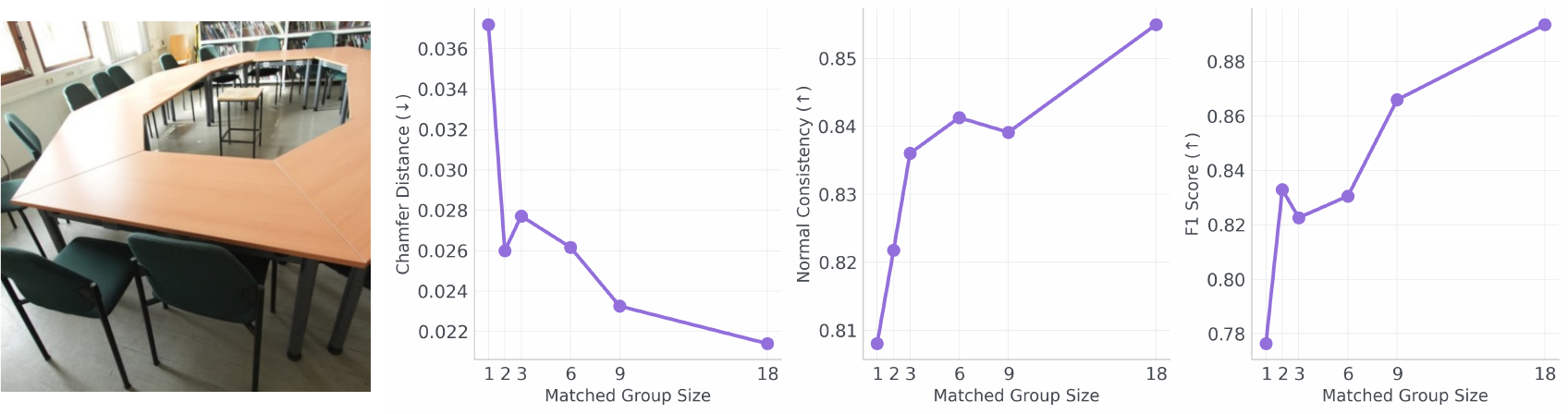}
\caption{
Reconstruction metrics for 18 chairs from ScanNet++ \textit{Scan 4} as the size of the group for joint reconstruction increases. 
} 
\label{fig:groupsize}
\end{figure}
\begin{figure}[t]
\centering
\includegraphics[width=\linewidth]{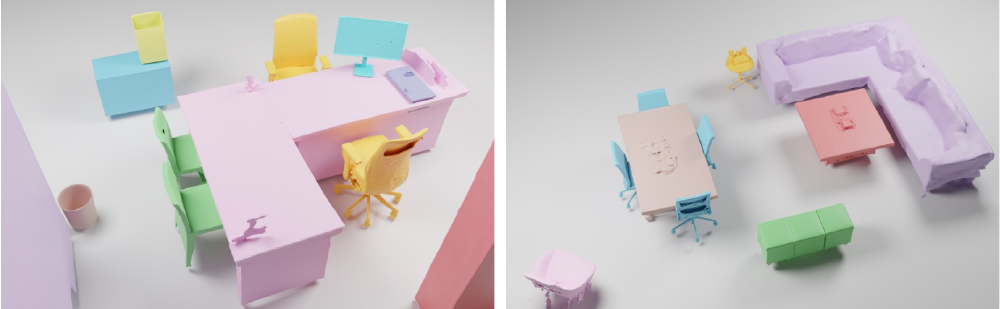}
\caption{
Reconstruction with all objects in a single group. 
} 
\label{fig:allobjs}
\end{figure}
\begin{figure*}
\centering
\includegraphics[width=\linewidth]{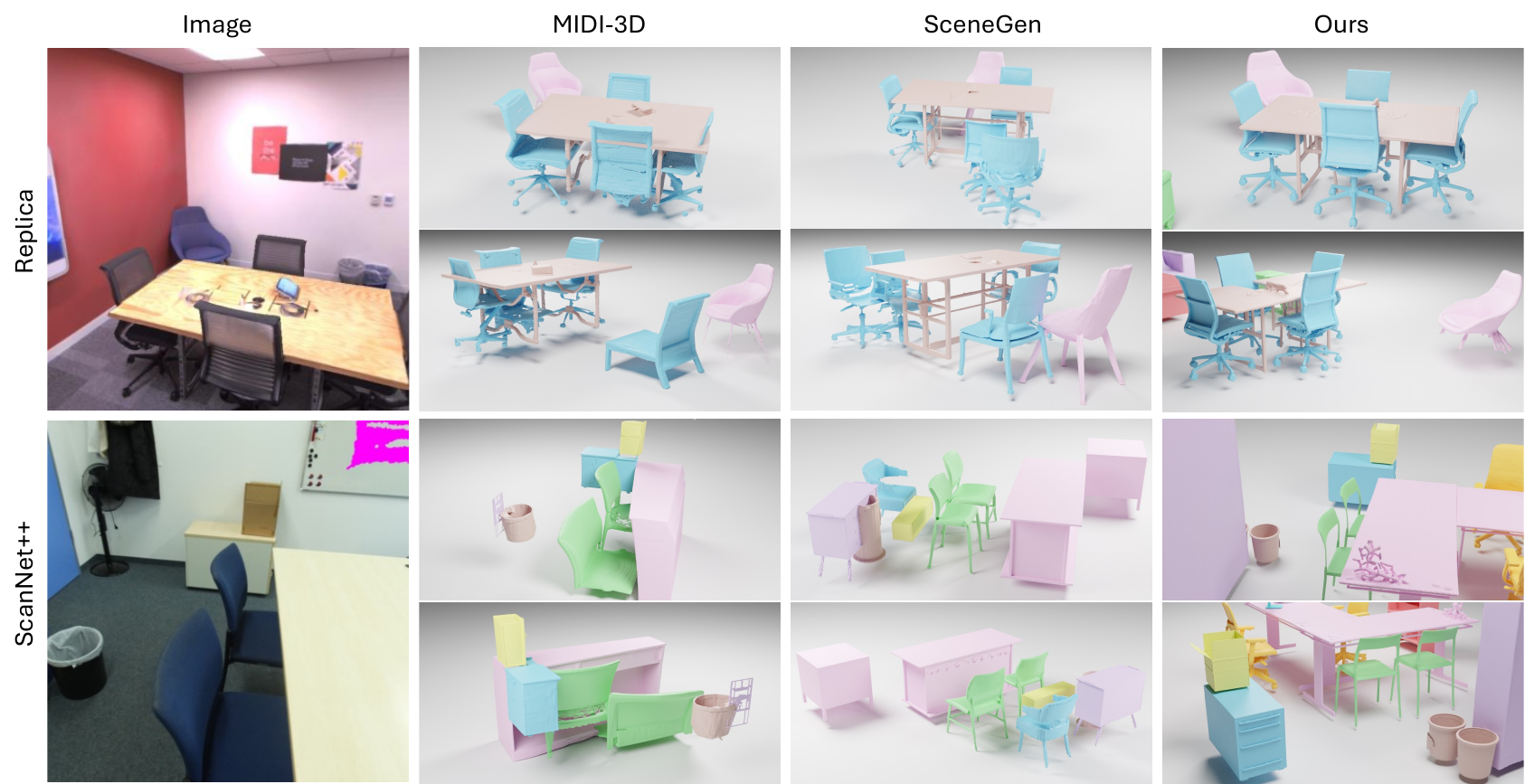}
\caption{
Comparison with MIDI-3D and SceneGen, both are conditioned on the single image displayed. Best viewed zoomed.
} 
\label{fig:img2scene}
\end{figure*}
\begin{figure*}
\centering
\includegraphics[width=\linewidth]{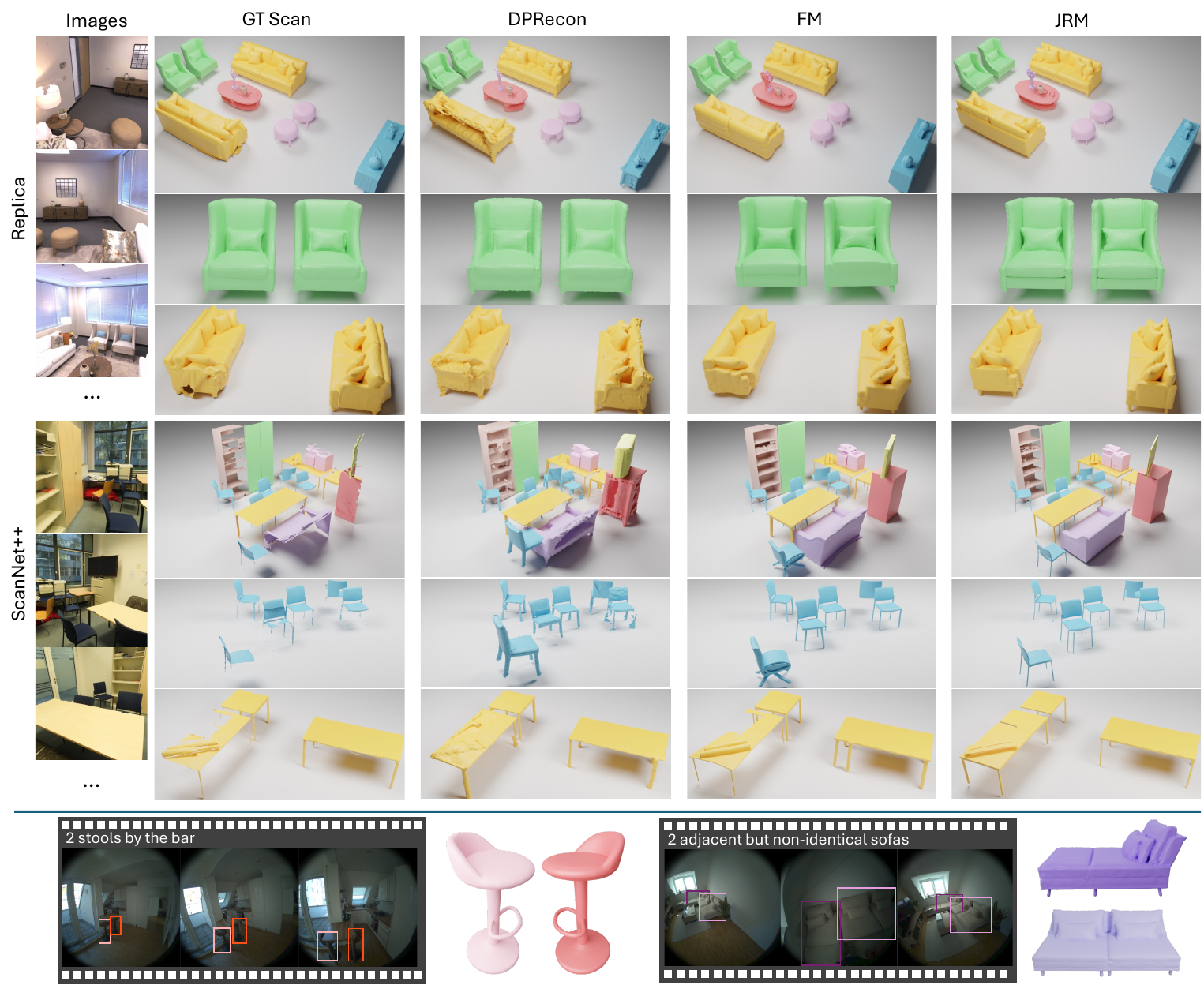}
\caption{
Additional qualitative results on real-world scenes, including ScanNet++ and Replica.
We use the same color for objects within the same match group. For each scene, we produce two close-up views focusing on the matched objects. 
} 
\label{fig:realscenes-supp}
\end{figure*}

\section{Model Architecture}
\label{sec:supp-method}

\subsection{ShapeR: Robust Conditional 3D Shape Generation from Casual Captures}


ShapeR~\cite{siddiqui2026shaper} uses multiple egocentric input modalities and robust training strategies to achieve object-centric 3D reconstruction from image sequences. In this section, we briefly summarize how the inputs are processed.

Given a scan of an environment, an off-the-shelf visual-inertial SLAM technique~\cite{engel2023project} is used to extract a sparse 3D pointcloud and camera poses. Subsequently, object instances are identified using a 3D instance detection approach~\cite{straub2024efm3d}.

For each detected object, its sparse points, corresponding image crops, 2D point mask projected on images, and text prompt generated by a vision-language model~\cite{meta2025llama4} are extracted. Camera poses are embedded as Pl\"{u}cker ray encodings and further concatenated with image tokens.

These multimodal conditions guide a 3D rectified flow matching model, which denoises a latent VecSet using a mixed architecture of single- and double-stream blocks, and decodes it to produce the 3D meshes. 

During training, ShapeR applies extensive augmentations to all modalities to simulate noisy and realistic inputs that further improve robustness. 
For images, the augmentations contain background compositing, occlusion overlays, visibility fog, resolution degradation, and photometric perturbations. 
For SLAM points, the augmentations simulate partial trajectories, a diverse range of point dropout strategies, Gaussian noise, and point occlusion.

It also leverages curriculum learning to ensure the robustness in real-world scenarios by employing two-stage training on object-level and scene-level data respectively.
We refer the full details of ShapeR to a separate material, \textsc{ShapeR.pdf}.

\subsection{Implementation Details.}

Our model extends a base ShapeR architecture~\cite{siddiqui2026shaper}. To make experimentation more feasible, we use a smaller variant of the ShapeR model as the basis for our experiments in this paper.
This variant of ShapeR, \fm, comprises 8 dual-stream and 16 single-stream blocks, each with 16 attention heads and a hidden width of 1024. It is trained for 300K steps on 128 NVIDIA H100 GPUs with an effective batch size of 1,536, progressively increasing the latent sequence length from 256 to 4,096.

The 3D VAE consists of an encoder of 8 transformer layers and a decoder of 16 layers, each with a hidden width of 768 and 12 attention heads. The VAE is trained for 200K steps with an effective batch size of 640 across 64 NVIDIA H100 GPUs.

We extend this architecture into a joint reconstruction model (\ours) by introducing coupled fusion blocks that alternate between single-stream blocks. We illustrate how the coupled attention mechanism builds upon a single-stream block in~\cref{fig:coupled}. Both \ours and \fm use identical encoders for multimodal inputs.

The \ours is initialized from the pretrained checkpoint of \fm. For the implementation of \ours, we offer two approaches: \emph{replace} every other pretrained single-stream block with a fresh coupled fusion block, or \emph{insert} a new coupled fusion block following each single-stream block. The latter essentially enlarges the model capacity of the base ShapeR. We compare their performance differences in~\cref{jrm_variants}.
A complete pipeline of applying \ours to the raw video input is illustrated in~\cref{fig:flowchart}.

\ours is trained using object pairs constructed from 80K 3D shapes. The initial learning rate is set at $1\times10^{-5}$, with a multi-step scheduler reducing it by 0.1 at 80K steps and by 0.05 at 160K steps. Training is conducted on 64 NVIDIA H100 GPUs with an effective batch size of 256, over 200K steps.
Again for feasibility of experiments, we don't perform curriculum learning that further trains ShapeR on the scene-level data.

\section{Data Preparation}
\label{sec:supp-data}

\subsection{Training Data}
For each training object, we rescale it diagonally within a unit cube and place the center at the origin.
We first sample a varied number of controlled viewpoints determined by the sampled radius of camera positions, calculated as $r_{cam}=\text{max}(\text{bbox}_{obj}) + \delta_{r}$, where $\delta_{r}$ is a random increment within the range of 0.5 to 1. By default, the camera is oriented towards the origin.
With fixed viewpoints, we then interpolate between viewpoints to produce a flying camera trajectory from a sequence of 100 camera viewpoints in total to render synthetic video capture. 

With a short video clip produced, we can obtain sparse SLAM points~\cite{engel2023project} and extract video frames with strong data augmentation applied as our conditions. We use 2 randomly sampled frames of resolution $224 \times 224$ as the image condition for each object. In particular, sparse structured SLAM points better simulate real-world noisy points and generalize well to more dense depth-back-projected points. 

\subsection{Evaluation Benchmarks}
We follow a heuristic algorithm proposed in~\citet{wu2024generalizing} to construct the synthetic benchmarks.
A synthetic scene is constructed by iteratively placing a new object to an existing arrangement that mimics real-world occlusion and clutter.
Examples of synthetic scenes are shown in~\cref{fig:syn_examples}.

\mypara{Scene layout generation.}
We arrange objects by sequentially adding new 3D shapes to the existing layout, ensuring that their 2D bounding boxes projected on the top-down view do not intersect with those of previously placed shapes. The goal is to create a scene where objects are near enough to present occlusion in rendering but do not overlap (see~\cref{fig:heuristics}).

For each 3D shape $S_i$ to be placed, we normalize its scale within a unit cube and randomly rotate it around the vertical axis. If no shapes have yet been placed, its starting position $\pmb{p}_i^0$ is set at the origin; otherwise, it is positioned at the mean location of the shapes placed previously.
A unit vector $\pmb{v}_i$ is randomly sampled for the direction of placement. The initial placement distance $d_i^0$ is calculated by summing the short sides of all positioned objects. Thus, the location of the new shape is determined by $d_i^0 \cdot \pmb{v}_i + \pmb{p}_i^0$.

In cases where the 2D bounding boxes of existing objects intersect with a new object, we increment $d_i^0$ by 0.05 iteratively until, after $N$ iterations, no intersections remain.
The final shape position is calculated as $(d_i^0 + 0.05 \times N) \cdot \pmb{v}_i + \pmb{p}_i^0$. 
The vertical position of the shape is also calculated with a random deviation sampled from $[-0.5,0.2]$. 

We repeat the procedure until all shapes are placed in the scene. Note that we only consider intersections among 2D bounding boxes from the top-down view as a simplification instead of using more expensive physics-based collision checks.

\mypara{Temporal instance repetition.}
To investigate the setting of temporal repetition of instances, we adopt the configuration setting of synthetic scenes from LivingScenes~\cite{zhu2023living} where a living scene is considered as a changing 3D environment of multiple objects dynamically transformed over time. 

We randomly sample a set of 3D shapes to build a scene from 6 categories, including ``chair'', ``table'', ``sofa'', ``bed'', ``pillow'' and ``lamp''. Each scene contains a varied number of objects, ranging from 4 to 8.
For the same set of selected objects, we generate 4 distinct scene arrangements using the scene layout generation approach described above.
One of the arrangements is considered the target reconstruction scene, while the other three arrangements simulate temporal changes of the target scene with rigid transformations at irregular intervals.

\mypara{Spatial instance repetition.}
In the setting of spatial instance repetition, we only generate one possible scene arrangement for the selected objects. 

We randomly sample a target object from 6 main categories, including ``chair'', ``table'', ``sofa``, ``bed'', ``storage'' and ``others''. The broad ``others'' category further contains 10 sub-categories with fewer instances each, including ``flower pot'', ``vase'', ``fan'', ``lamp'', ``tent'', ``ladder'', ``pillow'', ``cart'', ``office appliance'', and ``exercise weight''.
We intentionally choose three distinct source objects for the target: one identical, one similar, and one negative.

In a manner akin to generating pair-wise training data, a similar instance is randomly selected using pre-computed DuoDuoCLIP~\cite{lee2025duoduo} embeddings, ensuring the cosine similarity with the target object is greater than 0.65 and belongs to the same semantic category. A negative instance is randomly selected from different semantic categories.
Two more objects are randomly sampled from all available categories as occluders to create occlusion in renderings.

\mypara{Articulation.}
To study the generalization of \ours to dynamic articulated objects, we generated synthetic scenes with repetitive instances of an articulated object spawned into a scene at different motion states. 
We primarily concentrate on articulated furniture items intended for storage or functionality, like cabinets, dressers, dishwashers, microwaves, \etc.

Procedural programs can swiftly create these objects by altering kinematic graphs at the part level and adjusting spatial parameters like positions and sizes.
Initially, we select an articulated object and create three variations, each exhibiting distinct part-level deformations by employing the rest state and two random articulation states.
As with spatial instance repetition, we introduce two random objects into the scene to act as occluders.

\mypara{Synthetic scene rendering.}
We utilize a camera trajectory generation method similar to that applied for object-centric training data. The key difference is that the camera moves around in the entire scene rather than focusing on a single object. Moreover, the camera is directed towards a changing lookat target rather than the fixed origin.
We use back-projected rendered depth points as the point condition to \ours for evaluation on all synthetic scene benchmarks.

\section{Additional Experiments}
\label{sec:supp-experiments}

\subsection{Additional Results on Temporal Instance Repetition}
In~\cref{tab:livingscenes}, we present additional quantitative results on temporal instance repetition with predicted object matching by MORE$^2$~\cite{zhu2023living}. This differs from Table 2 in the main paper, which used ground truth object matching instead of a predicted matching. Therefore, the results for ``No Rescan" remain the same as before, but change with further rescans where ideal object matching is no longer assumed.

For these experiments, we evaluate two variants of \fm and \ours. We train a version of \fm and \ours conditioning solely on the point condition, and build on a curated training dataset that matches the size of the MORE$^2$ training data, offering a fair comparison with MORE$^2$~\cite{zhu2023living}. 
We also present results of \fm and \ours trained on our full data and conditioned on three modalities. 

In these results, we see a continuation of the trends observed in the main paper. Even with both imperfect matching and alignment, \ours is able to improve reconstruction accuracy with addition source scans. However, the additional noise introduced by imperfect matching causes the baseline methods' reconstructions to deteriorate as the environment is re-observed. 

\subsection{Coupled Blocks Insertion Ablation} 
\label{jrm_variants}
We study how \ours performs with different implementations of the framework. In addition to \emph{replacing} every other single-stream block with a coupled fusion block in the original \fm architecture, we can instead \emph{insert} a coupled fusion block after each single-stream block, which results in a larger model capacity, 40 attention layers in total. 

The results in~\cref{tab:jrm_variants} show that both implementations fulfill the goal of observation fusion, but that additional model capacity yields further performance improvements across different paired objects.

\subsection{Additional Qualitative Results}

\mypara{Robustness to dissimilarity.} 
We clarify that JRM does not ``force" dissimilar objects to look alike, nor do we filter for near-identical training pairs. During training, the model is exposed to pairs of distinct objects with a probability of $0.1$, ablated in Paper Tab. 5. In~\cref{fig:similar}, we show JRM respects instance-specific identity; it correctly preserves unique features like throw cushions or specific leg geometries on one while omitting them from others. Paper Tab. 2 quantifies this robustness where ``similar" matches with CLIP scores between 0.65 and 0.9. The degradation from non-identical matches is significantly less severe than the baseline, showing JRM balances shared features between objects rather than blindly collapsing them.

\mypara{Implicit Segmentation.} A common failure mode that we observe in our experiments is the generation of additional geometry that does not belong to the target object. We refer to these cases as failures of implicit segmentation, as the reconstruction method is unable to isolate the target from distractor content, \eg from occluding objects, in the conditioning inputs. 

We show in~\cref{fig:segment} that joint reconstruction with implicit observation fusion also improves object segmentation during geometry generation. 
In the first case, taking extra unaligned source observations helps to eliminate artifacts due to a restricted perspective of the target observation.
In the second case, simply combining observations using explicit alignment can result in worse reconstruction. However, joint reconstruction with implicit aggregation is capable of avoiding such performance degradation.

\mypara{Generalisation to larger group sizes.} 
We investigate the generalisation of JRM to larger batches of joint reconstruction, from pairs, as seen during training to larger groups. Specifically, we use \textit{Scan 4} from ScanNet++, which has 18 similar chairs. \Cref{fig:groupsize} shows chair reconstruction accuracy. Although trained on pairs, metrics consistently improve as the size of the joint group increases.

\mypara{Comparison with MIDI-3D and SceneGen.} 
We perform qualitative comparisons on Replica and ScanNet++ in ~\cref{fig:img2scene}. While these prior works are conditioned on a single image, JRM leverages multi-view inputs; ground-truth segmentation is provided to all methods. 
Some baseline errors, such as pose and scale inaccuracies, may stem from the monocular input. However, more fundamental issues exhibited by both MIDI-3D and SceneGen, \eg implausible geometry and inconsistent styles, are largely avoided by JRM. We attribute this to the significantly larger scale of data accessible by our object-centric training approach compared to scene-limited alternatives.

\mypara{Full-scene group.} 
Using the same scenes as~\cref{fig:img2scene}, we evaluate a baseline with all objects in a single group in~\cref{fig:allobjs}. There is a regression compared to class-based groups, but the reconstructions remain representative. 

\mypara{Real-world Scenes.} We present extra qualitative results on real-world scenes in~\cref{fig:realscenes-supp}, including sources of Replica~\cite{straub2019replica}, ScanNet++~\cite{yeshwanth2023scannet++} and a real apartment scanned using Aria glasses~\cite{engel2023project}.

\end{document}